\documentclass[10pt,twocolumn,letterpaper]{article}

\usepackage{iccv}
\usepackage{times}
\usepackage{epsfig}
\usepackage{graphicx}
\usepackage{amsmath}
\usepackage{amssymb}
\usepackage{xcolor}
\usepackage{subcaption}
\usepackage{dblfloatfix}
\usepackage{setspace}
\usepackage{bbm}

\usepackage[pagebackref=false,breaklinks=true,letterpaper=true,colorlinks=true,bookmarks=false]{hyperref}

\iccvfinalcopy 

\def\iccvPaperID{****} 

\usepackage{gensymb}
\usepackage{booktabs}
\usepackage{cite}

\definecolor{skyblue3}{RGB}{ 32,  74, 135}
\definecolor{scarletred3}{RGB}{164,   0,   0}
\definecolor{aluminium3}{RGB}{186, 189, 182}
\definecolor{aluminium5}{RGB}{ 85,  87,  83}
\definecolor{chameleon3}{RGB}{ 78, 154,   6}

\ificcvfinal\fi
\begin{document}

\title{Objects as Points}

\author{Xingyi Zhou\\
UT Austin\\
{\tt\small zhouxy@cs.utexas.edu}
\and
Dequan Wang\\
UC Berkeley\\
{\tt\small dqwang@cs.berkeley.edu}
\and
Philipp Kr\"ahenb\"uhl\\
UT Austin\\
{\tt\small philkr@cs.utexas.edu}
}

\maketitle

\begin{abstract}
Detection identifies objects as axis-aligned boxes in an image.
Most successful object detectors enumerate a nearly exhaustive list of potential object locations and classify each.
This is wasteful, inefficient, and requires additional post-processing.
In this paper, we take a different approach.
We model an object as a single point --- the center point of its bounding box.
Our detector uses keypoint estimation to find center points and regresses to all other object properties, such as size, 3D location, orientation, and even pose.
Our center point based approach, CenterNet, is end-to-end differentiable, simpler, faster, and more accurate than corresponding bounding box based detectors.
CenterNet achieves the best speed-accuracy trade-off on the MS COCO dataset, with $28.1\%$ AP at 142 FPS, $37.4\%$ AP at 52 FPS, and $45.1\%$ AP with multi-scale testing at 1.4 FPS.
We use the same approach to estimate 3D bounding box in the KITTI benchmark and human pose on the COCO keypoint dataset.
Our method performs competitively with sophisticated multi-stage methods and runs in real-time.

\end{abstract}

\section{Introduction}

Object detection powers many vision tasks like instance segmentation~\cite{he2017mask,li2017fully,chen2018deeplab}, pose estimation~\cite{cao2018openpose,fang2017rmpe,newell2017associative}, tracking~\cite{kalal2012tracking,hu2018joint}, and action recognition~\cite{carreira2017quo}.
It has down-stream applications in surveillance~\cite{xu2018attention}, autonomous driving~\cite{wang2018deep}, and visual question answering~\cite{antol2015vqa}.
Current object detectors represent each object through an axis-aligned bounding box that tightly encompasses the object~\cite{girshick2014rich,girshick2015fast,ren2015faster,redmon2016you,lin2018focal}.
They then reduce object detection to image classification of an extensive number of potential object bounding boxes.
For each bounding box, the classifier determines if the image content is a specific object or background.
One-stage detectors~\cite{redmon2016you,lin2018focal} slide a complex arrangement of possible bounding boxes, called anchors, over the image and classify them directly without specifying the box content.
Two-stage detectors~\cite{girshick2014rich,girshick2015fast,ren2015faster} recompute image-features for each potential box, then classify those features.
Post-processing, namely non-maxima suppression, then removes duplicated detections for the same instance by computing bounding box IoU.
This post-processing is hard to differentiate and train~\cite{hosang2017learning}, hence most current detectors are not end-to-end trainable.
Nonetheless, over the past five years~\cite{girshick2014rich}, this idea has achieved good empirical success~\cite{he2017mask,xie2017aggregated,Dai_2017_ICCV,singh2018analysis,huang2017speed,Jiang_2018_ECCV,liu2018path,zhu2018deformable,sniper2018,li2019scale,zhu2019feature}.
Sliding window based object detectors are however a bit wasteful, as they need to enumerate all possible object locations and dimensions.

\begin{figure}[t]
\centering
   \vspace{-1em}
   \includegraphics[width=0.95\linewidth]{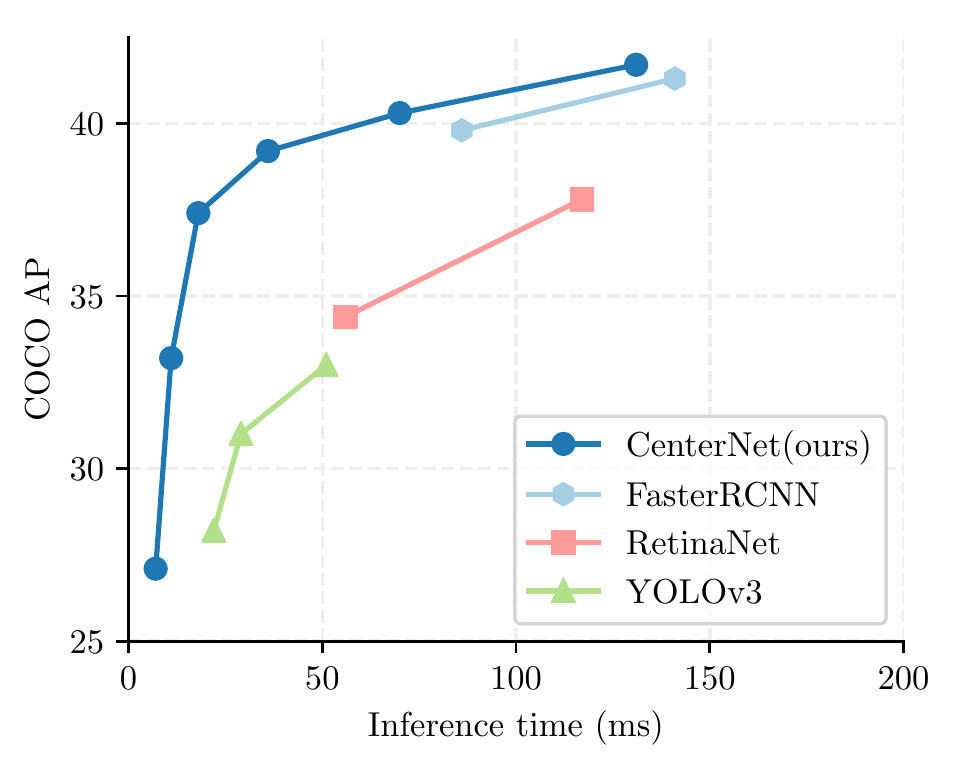}
   \vspace{-1em}
   \caption{Speed-accuracy trade-off on COCO validation for real-time detectors. The proposed CenterNet outperforms a range of state-of-the-art algorithms.}
   \vspace{-1em}
\label{fig:tradeoff}
\label{fig:onecol}
\end{figure}


\begin{figure*}[t]
\centering
   \includegraphics[width=0.95\linewidth]{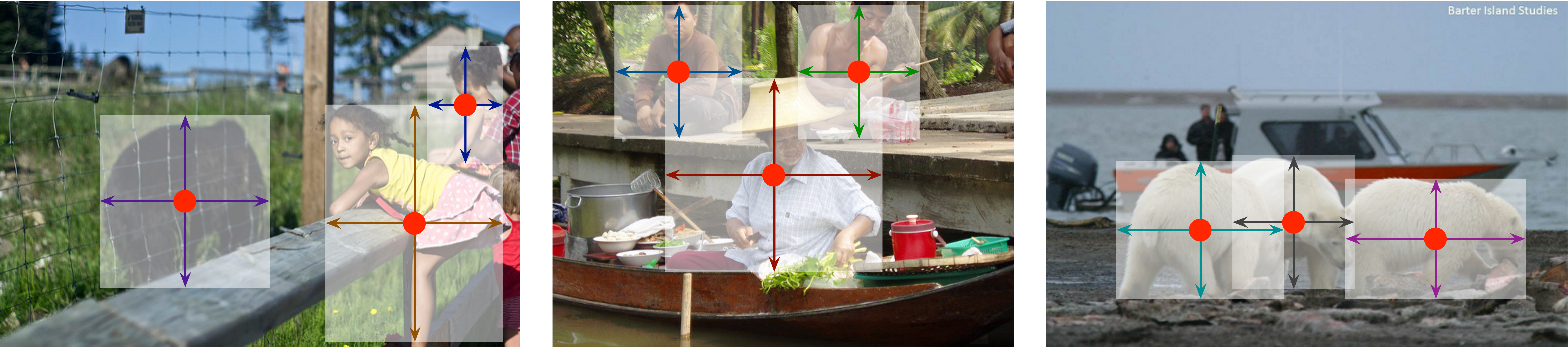}
    \vspace{-1mm}
   \caption{We model an object as the center point of its bounding box. The bounding box size and other object properties are inferred from the keypoint feature at the center. Best viewed in color.}
   \vspace{-1.0em}
\label{fig:teaser}
\end{figure*}

In this paper, we provide a much simpler and more efficient alternative.
We represent objects by a single point at their bounding box center (see Figure ~\ref{fig:teaser}).
Other properties, such as object size, dimension, 3D extent, orientation, and pose are then regressed directly from image features at the center location.
Object detection is then a standard keypoint estimation problem~\cite{cao2018openpose,newell2017associative,zhou2018starmap}.
We simply feed the input image to a fully convolutional network~\cite{long2015fully,newell2016stacked} that generates a heatmap.
Peaks in this heatmap correspond to object centers.
Image features at each peak predict the objects bounding box height and weight.
The model trains using standard dense supervised learning~\cite{newell2017associative,zhou2018starmap}.
Inference is a single network forward-pass, without non-maximal suppression for post-processing.

Our method is general and can be extended to other tasks with minor effort.
We provide experiments on 3D object detection~\cite{Geiger2012CVPR} and multi-person human pose estimation~\cite{cao2017realtime}, by predicting additional outputs at each center point (see Figure~\ref{fig:head}).
For 3D bounding box estimation, we regress to the object absolute depth, 3D bounding box dimensions, and object orientation~\cite{mousavian20173d}.
For human pose estimation, we consider the 2D joint locations as offsets from the center and directly regress to them at the center point location.

The simplicity of our method, CenterNet, allows it to run at a very high speed (Figure~\ref{fig:tradeoff}).
With a simple Resnet-18 and up-convolutional layers~\cite{xiao2018simple}, our network runs at 142 FPS with $28.1\%$ COCO bounding box AP. 
With a carefully designed keypoint detection network, DLA-34~\cite{yu2018deep}, our network achieves $37.4\%$ COCO AP at 52 FPS.
Equipped with the state-of-the-art keypoint estimation network, Hourglass-104~\cite{Law_2018_ECCV,newell2016stacked}, and multi-scale testing, our network achieves $45.1\%$ COCO AP at 1.4 FPS.
On 3D bounding box estimation and human pose estimation, we perform competitively with state-of-the-art at a higher inference speed.
Code is available at \url{https://github.com/xingyizhou/CenterNet}.

\section{Related work}

\paragraph{Object detection by region classification.}
One of the first successful deep object detectors, RCNN~\cite{girshick2014rich}, enumerates object location from a large set of region candidates~\cite{uijlings2013selective}, crops them, and classifies each using a deep network.
Fast-RCNN~\cite{girshick2015fast} crops image features instead, to save computation. 
However, both methods rely on slow low-level region proposal methods.

\paragraph{Object detection with implicit anchors.}
 
Faster RCNN~\cite{ren2015faster} generates region proposal within the detection network. 
It samples fixed-shape bounding boxes (anchors) around a low-resolution image grid and classifies each into ``foreground or not''.
An anchor is labeled foreground with a $>\!\!\!0.7$ overlap with any ground truth object, background with a $<\!\!\!0.3$ overlap, or ignored otherwise.
Each generated region proposal is again classified~\cite{girshick2015fast}.
Changing the proposal classifier to a multi-class classification forms the basis of one-stage detectors.
Several improvements to one-stage detectors include anchor shape priors~\cite{redmon2017yolo9000,redmon2018yolov3}, different feature resolution~\cite{liu2016ssd}, and loss re-weighting among different samples~\cite{lin2018focal}.

\begin{figure}[t]
\begin{subfigure}[t]{0.53\linewidth}
  \centering
   \includegraphics[width=\linewidth,trim=0 0 0 0,clip]{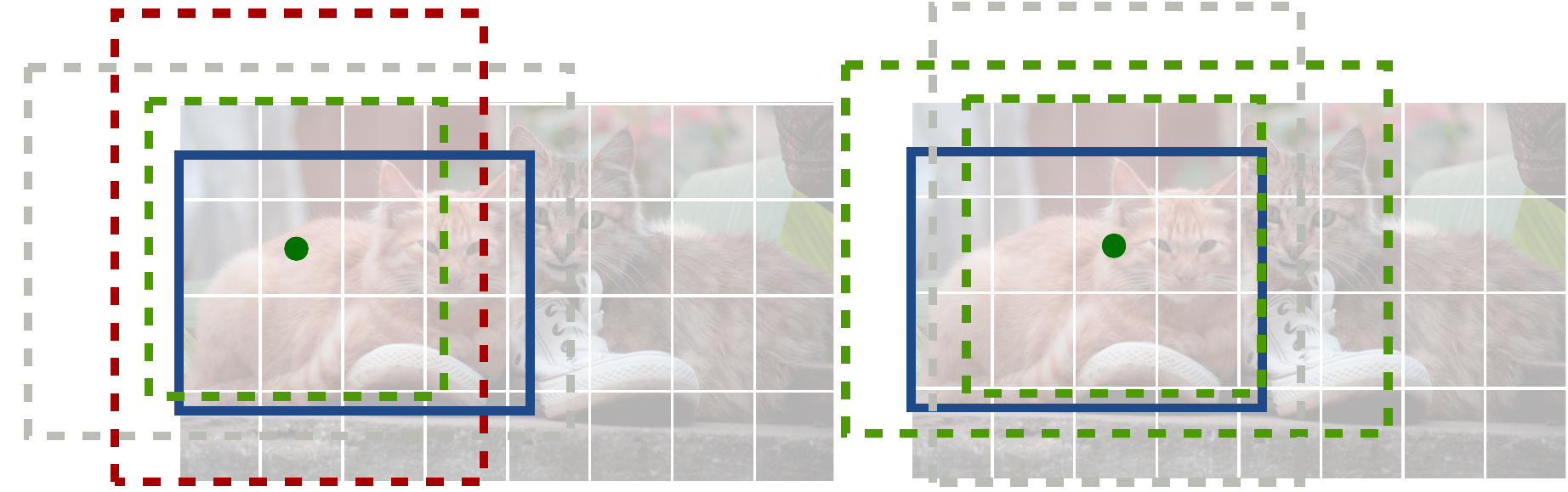}
   \caption{Standard anchor based detection. Anchors count as \textcolor{chameleon3}{positive} with an overlap $IoU>0.7$ to any \textcolor{skyblue3}{object}, \textcolor{scarletred3}{negative} with an overlap $IoU<0.3$, or are \textcolor{aluminium3}{ignored} otherwise.}
\end{subfigure}
\hspace{1mm}
\begin{subfigure}[t]{0.43\linewidth}
  \centering
   \includegraphics[width=0.8\linewidth]{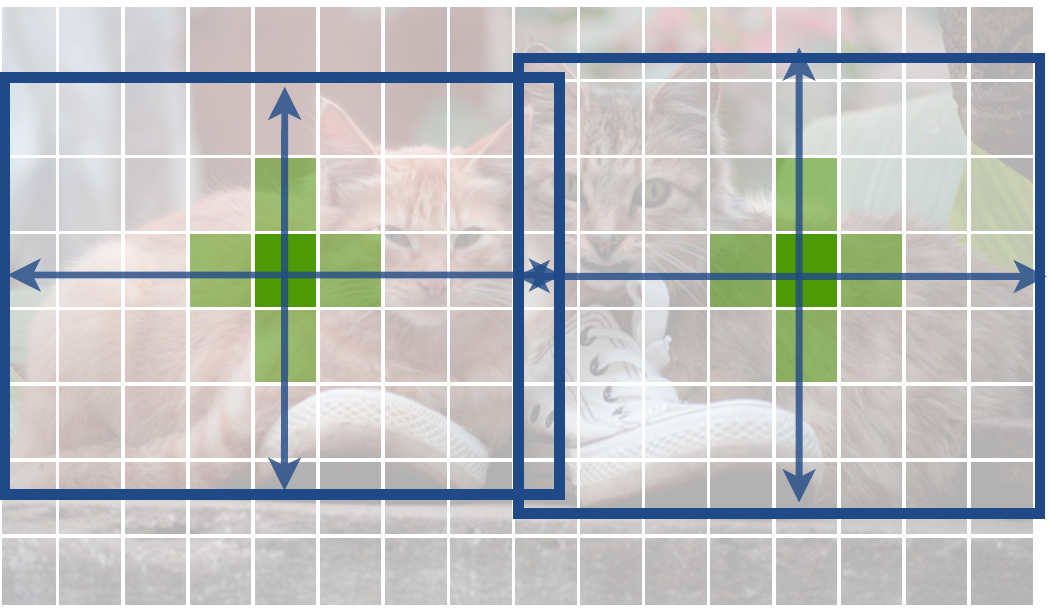}
   \caption{Center point based detection. The \textcolor{chameleon3}{center pixel} is assigned to the \textcolor{skyblue3}{object}. Nearby points have a reduced negative loss. Object size is regressed.}
\end{subfigure}
   \caption{Different between anchor-based detectors (a) and our center point detector (b). Best viewed on screen.}
\label{fig:anchor}
\end{figure}

Our approach is closely related to anchor-based one-stage approaches~\cite{redmon2016you,liu2016ssd,lin2018focal}.
A center point can be seen as a single shape-agnostic anchor (see Figure~\ref{fig:anchor}).
However, there are a few important differences.
First, our CenterNet assigns the ``anchor'' based solely on location, not box overlap~\cite{girshick2015fast}.
We have no manual thresholds~\cite{girshick2015fast} for foreground and background classification.
Second, we only have one positive ``anchor'' per object, and hence do not need Non-Maximum Suppression (NMS)~\cite{bodla2017soft}.
We simply extract local peaks in the keypoint heatmap~\cite{newell2017associative,cao2017realtime}.
Third, CenterNet uses a larger output resolution (output stride of $4$) compared to traditional object detectors~\cite{he2016deep,he2017mask} (output stride of $16$).
This eliminates the need for multiple anchors~\cite{singh2018analysis}.

\paragraph{Object detection by keypoint estimation.}
We are not the first to use keypoint estimation for object detection.
CornerNet~\cite{Law_2018_ECCV} detects two bounding box corners as keypoints, while ExtremeNet~\cite{zhou2019bottomup} detects the top-, left-, bottom-, right-most, and center points of all objects.
Both these methods build on the same robust keypoint estimation network as our CenterNet.
However, they require a combinatorial grouping stage after keypoint detection, which significantly slows down each algorithm.
Our CenterNet, on the other hand, simply extracts a single center point per object without the need for grouping or post-processing.

\paragraph{Monocular 3D object detection.}
3D bounding box estimation powers autonomous driving~\cite{Geiger2012CVPR}.
Deep3Dbox~\cite{mousavian20173d} uses a slow-RCNN~\cite{girshick2014rich} style framework, by first detecting 2D objects~\cite{ren2015faster} and then feeding each object into a 3D estimation network.
3D RCNN~\cite{3DRCNN_CVPR18} adds an additional head to Faster-RCNN~\cite{ren2015faster} followed by a 3D projection.
Deep Manta~\cite{chabot2017deep} uses a coarse-to-fine Faster-RCNN~\cite{ren2015faster} trained on many tasks.
Our method is similar to a one-stage version of Deep3Dbox~\cite{mousavian20173d} or 3DRCNN~\cite{3DRCNN_CVPR18}.
As such, CenterNet is much simpler and faster than competing methods.

\section{Preliminary}
\label{sec:prelim}
Let $I \in R^{W \times H \times 3}$ be an input image of width $W$ and height $H$.
Our aim is to produce a keypoint heatmap $\hat Y \in [0,1]^{\frac{W}{R} \times \frac{H}{R} \times C}$, where $R$ is the output stride and $C$ is the number of keypoint types.
Keypoint types include $C=17$ human joints in human pose estimation~\cite{cao2017realtime,xiao2018simple}, or $C=80$ object categories in object detection~\cite{Law_2018_ECCV,zhou2019bottomup}.
We use the default output stride of $R=4$ in literature~\cite{newell2016stacked,cao2017realtime,papandreou2017towards}.
The output stride downsamples the output prediction by a factor $R$.
A prediction $\hat Y_{x,y,c} = 1$ corresponds to a detected keypoint, while $\hat Y_{x,y,c} = 0$ is background.
We use several different fully-convolutional encoder-decoder networks to predict $\hat Y$ from an image $I$: A stacked hourglass network~\cite{newell2016stacked,Law_2018_ECCV}, up-convolutional residual networks (ResNet)~\cite{he2016deep,xiao2018simple}, and deep layer aggregation (DLA)~\cite{yu2018deep}.

We train the keypoint prediction network following Law and Deng~\cite{Law_2018_ECCV}.
For each ground truth keypoint ${p \in \mathcal{R}^2}$ of class $c$, we compute a low-resolution equivalent ${\tilde p = \lfloor \frac{p}{R} \rfloor}$.
We then splat all ground truth keypoints onto a heatmap ${Y \in [0,1]^{\frac{W}{R} \times \frac{H}{R} \times C}}$ using a Gaussian kernel
${Y_{xyc} = \exp\left(-\frac{(x-\tilde p_x)^2+(y-\tilde p_y)^2}{2\sigma_p^2}\right)}$,
where $\sigma_p$ is an object size-adaptive standard deviation~\cite{Law_2018_ECCV}.
If two Gaussians of the same class overlap, we take the element-wise maximum~\cite{cao2017realtime}.
The training objective is a penalty-reduced pixel-wise logistic regression with focal loss~\cite{lin2018focal}:
\begin{equation}
    L_k = \frac{-1}{N} \sum_{xyc}
    \begin{cases}
        (1 - \hat{Y}_{xyc})^{\alpha} 
        \log(\hat{Y}_{xyc}) & \!\text{if}\ Y_{xyc}=1\vspace{2mm}\\
        \begin{array}{c}
        (1-Y_{xyc})^{\beta} 
        (\hat{Y}_{xyc})^{\alpha}\\
        \log(1-\hat{Y}_{xyc})
        \end{array}
        & \!\text{otherwise}
    \end{cases}
\end{equation}
where $\alpha$ and $\beta$ are hyper-parameters of the focal loss~\cite{lin2018focal}, and $N$ is the number of keypoints in image $I$.
The normalization by $N$ is chosen as to normalize all positive focal loss instances to $1$.
We use $\alpha=2$ and $\beta=4$ in all our experiments, following Law and Deng~\cite{Law_2018_ECCV}.

To recover the discretization error caused by the output stride, we additionally predict a local offset $\hat O \in \mathcal{R}^{\frac{W}{R} \times \frac{H}{R} \times 2}$ for each center point.
All classes $c$ share the same offset prediction.
The offset is trained with an L1 loss
\begin{equation}
    L_{off} = \frac{1}{N}\sum_{p} \left|\hat O_{\tilde p} - \left(\frac{p}{R} - \tilde p\right)\right|\label{eq:off_loss}.
\end{equation}
The supervision acts only at keypoints locations $\tilde p$, all other locations are ignored.

In the next section, we will show how to extend this keypoint estimator to a general purpose object detector.

\section{Objects as Points}
\label{sec:technic}
\begin{figure}[t]
  \begin{subfigure}[t]{0.33\linewidth}
   \centering
   \includegraphics[width=0.95\linewidth,page=1]{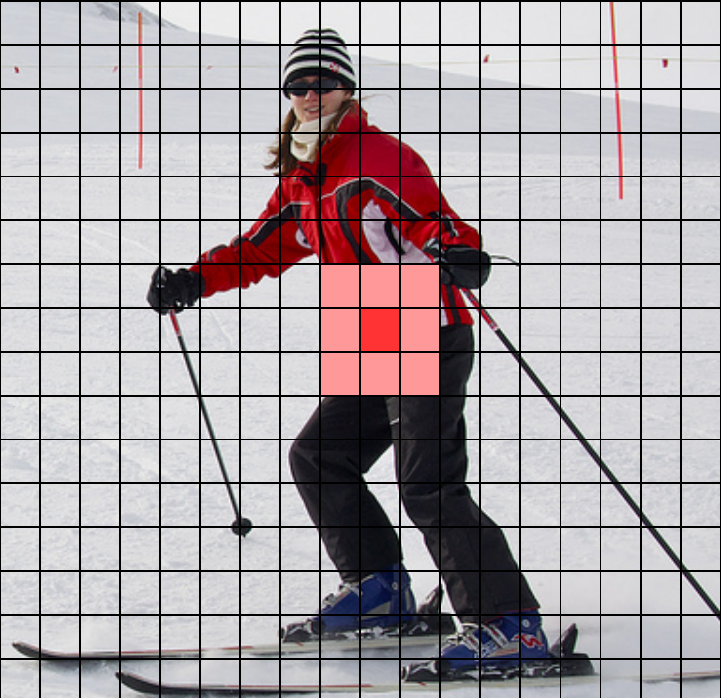}
   \caption*{keypoint heatmap [C]}
  \end{subfigure}%
  \begin{subfigure}[t]{0.33\linewidth}
   \centering
   \includegraphics[width=0.95\linewidth,page=3]{figs/head2.pdf}
   \caption*{local offset [2]}
  \end{subfigure}%
  \begin{subfigure}[t]{0.33\linewidth}
   \centering
   \includegraphics[width=0.95\linewidth,page=2]{figs/head2.pdf}
   \caption*{object size [2]}
  \end{subfigure}
  
  \begin{subfigure}[t]{0.33\linewidth}
   \centering
   \includegraphics[width=0.95\linewidth,page=4]{figs/head2.pdf}
   \caption*{3D size [3]}
  \end{subfigure}%
  \begin{subfigure}[t]{0.33\linewidth}
   \centering
   \includegraphics[width=0.95\linewidth,page=5]{figs/head2.pdf}
   \caption*{depth [1]}
  \end{subfigure}%
  \begin{subfigure}[t]{0.33\linewidth}
   \centering
   \includegraphics[width=0.95\linewidth,page=6]{figs/head2.pdf}
   \caption*{orientation [8]}
  \end{subfigure}

  \begin{subfigure}[t]{0.33\linewidth}
   \centering
   \includegraphics[width=0.95\linewidth,page=7]{figs/head2.pdf}
   \caption*{joint locations [$k\times 2$]}
  \end{subfigure}%
  \begin{subfigure}[t]{0.33\linewidth}
   \centering
   \includegraphics[width=0.95\linewidth,page=8]{figs/head2.pdf}
   \caption*{joint heatmap [$k$]}
  \end{subfigure}%
  \begin{subfigure}[t]{0.33\linewidth}
   \centering
   \includegraphics[width=0.95\linewidth,page=9]{figs/head2.pdf}
   \caption*{joint offset [$2$]}
  \end{subfigure}
    \vspace{-2.5mm}
   \caption{Outputs of our network for different tasks: \emph{top} for object detection, \emph{middle} for 3D object detection, \emph{bottom:} for pose estimation. All modalities are produced from a common backbone, with a different $3 \times 3$ and $1 \times 1$ output convolutions separated by a ReLU. The number in brackets indicates the output channels. See section~\ref{sec:technic} for details.}
\label{fig:head}
\end{figure}

Let $(x_1^{(k)}, y_1^{(k)}, x_2^{(k)}, y_2^{(k)})$ be the bounding box of object $k$ with category $c_k$.
Its center point is lies at $p_k = (\frac{x_1^{(k)} + x_2^{(k)}}{2}, \frac{y_1^{(k)} + y_2^{(k)}}{2})$.
We use our keypoint estimator $\hat Y$ to predict all center points.
In addition, we regress to the object size $s_k = (x_2^{(k)} - x_1^{(k)}, y_2^{(k)} - y_1^{(k)})$ for each object $k$.
To limit the computational burden, we use a single size prediction $\hat S  \in \mathcal{R}^{\frac{W}{R} \times \frac{H}{R} \times 2}$ for all object categories.
We use an L1 loss at the center point similar to Objective~\ref{eq:off_loss}:
\begin{equation}
    L_{size} = \frac{1}{N}\sum_{k=1}^{N} \left|\hat S_{p_k} - s_k\right|.
\label{eq:size_loss}
\end{equation}
We do not normalize the scale and directly use the raw pixel coordinates.
We instead scale the loss by a constant $\lambda_{size}$.
The overall training objective is
\begin{equation}
    L_{det} = L_{k} + \lambda_{size} L_{size} + \lambda_{off}L_{off}.
\label{eq:total_loss}
\end{equation}
We set $\lambda_{size} = 0.1$ and $ \lambda_{off} = 1$ in all our experiments unless specified otherwise.
We use a single network to predict the keypoints $\hat Y$, offset $\hat O$, and size $\hat S$.
The network predicts a total of $C + 4$ outputs at each location.
All outputs share a common fully-convolutional backbone network.
For each modality, the features of the backbone are then passed through a separate $3 \times 3$ convolution, ReLU and another $1 \times 1$ convolution.
Figure~\ref{fig:head} shows an overview of the network output.
Section~\ref{sec:networks} and supplementary material contain additional architectural details.

\paragraph{From points to bounding boxes}
At inference time, we first extract the peaks in the heatmap for each category independently.
We detect all responses whose value is greater or equal to its 8-connected neighbors and keep the top $100$ peaks.
Let $\hat{\mathcal{P}}_c$ be the set of $n$ detected center points $\hat{\mathcal{P}} = \{(\hat x_i, \hat y_i)\}_{i = 1}^{n}$ of class $c$.
Each keypoint location is given by an integer coordinates $(x_i,y_i)$.
We use the keypoint values $\hat Y_{x_iy_ic}$ as a measure of its detection confidence, and produce a bounding box at location
\begin{align*}
 (&\hat x_i + \delta \hat x_i - \hat w_i / 2,\ \ \hat y_i + \delta \hat y_i - \hat h_i / 2,\\
 &\hat x_i + \delta \hat x_i + \hat w_i / 2,\ \ \hat y_i + \delta \hat y_i + \hat h_i / 2),
\end{align*}
where $(\delta \hat x_i, \delta \hat y_i) = \hat O_{\hat x_i,\hat y_i}$ is the offset prediction and $(\hat w_i,\hat h_i) = \hat S_{\hat x_i,\hat y_i}$ is the size prediction.
All outputs are produced directly from the keypoint estimation without the need for IoU-based non-maxima suppression (NMS) or other post-processing.
The peak keypoint extraction serves as a sufficient NMS alternative and can be implemented efficiently on device using a $3 \times 3$ max pooling operation.

\subsection{3D detection}

3D detection estimates a three-dimensional bounding box per objects and requires three additional attributes per center point: depth, 3D dimension, and orientation.
We add a separate head for each of them.
The depth $d$ is a single scalar per center point.
However, depth is difficult to regress to directly.
We instead use the output transformation of Eigen \etal~\cite{eigen2014depth} and $d = 1 / \sigma(\hat d) - 1$, where $\sigma$ is the sigmoid function.
We compute the depth as an additional output channel $\hat D \in [0,1]^{\frac{W}{R} \times \frac{H}{R}}$ of our keypoint estimator.
It again uses two convolutional layers separated by a ReLU.
Unlike previous modalities, it uses the inverse sigmoidal transformation at the output layer.
We train the depth estimator using an L1 loss in the original depth domain, after the sigmoidal transformation.

The 3D dimensions of an object are three scalars.
We directly regress to their absolute values in meters using a separate head $\hat \Gamma \in \mathcal{R}^{\frac{W}{R} \times \frac{H}{R} \times 3}$ and an L1 loss.

Orientation is a single scalar by default.
However, it can be hard to regress to.
We follow Mousavian \etal~\cite{mousavian20173d} and represent the orientation as two bins with in-bin regression.
Specifically, the orientation is encoded using $8$ scalars, with $4$ scalars for each bin.
For one bin, two scalars are used for softmax classification and the rest two scalar regress to an angle within each bin.
Please see the supplementary for details about these losses.

\begin{table*}[ht]
\centering
\begin{tabular}{l c@{\ \ }c@{\ \ }c @{\qquad} c@{\ \ }c@{\ \ }c @{\qquad} c@{\ \ }c@{\ \ }c @{\qquad} c@{\ \ }c@{\ \ }c @{\qquad} c@{\ \ }c@{\ \ }c}
\hline
 &  \multicolumn{3}{c}{AP} & \multicolumn{3}{c}{$AP_{50}$}  & \multicolumn{3}{c}{$AP_{75}$} & \multicolumn{3}{c}{Time (ms)} & \multicolumn{3}{c}{FPS}\\
 & N.A. & F & MS & N.A. & F & MS & N.A. & F & MS & N.A. & F & MS & N.A. & F & MS\\
\hline
Hourglass-104 & \textbf{40.3} & \textbf{42.2} & \textbf{45.1} & \textbf{59.1} & \textbf{61.1} & \textbf{63.5} & \textbf{44.0} & \textbf{46.0} & \textbf{49.3} & 71 & 129 & 672 & 14 & 7.8 & 1.4\\
DLA-34 & 37.4 & 39.2 & 41.7  & 55.1 & 57.0 & 60.1  & 40.8 & 42.7 & 44.9  & 19 & 36 & 248 & 52 & 28 & 4\\
ResNet-101 & 34.6 & 36.2 & 39.3 & 53.0 & 54.8 & 58.5 & 36.9 & 38.7 & 42.0 & 22 & 40 & 259 & 45 & 25 & 4\\
ResNet-18 & 28.1 & 30.0 & 33.2 & 44.9 & 47.5 & 51.5 & 29.6 & 31.6 & 35.1 & \textbf{7} & \textbf{14} & \textbf{81} & \textbf{142} & \textbf{71} & \textbf{12}\\
\hline
\end{tabular}
\caption{Speed / accuracy trade off for different networks on COCO validation set. We show results without test augmentation (N.A.), flip testing (F), and multi-scale augmentation (MS).}
\label{tab:network}
\vspace{-1em}
\end{table*}

\subsection{Human pose estimation}

Human pose estimation aims to estimate $k$ 2D human joint locations for every human instance in the image ($k = 17$ for COCO).
We considered the pose as a $k \times 2$-dimensional property of the center point, and parametrize each keypoint by an offset to the center point.
We directly regress to the joint offsets (in pixels) $\hat J \in \mathcal{R}^{\frac{W}{R} \times \frac{H}{R} \times k \times 2}$ with an L1 loss.
We ignore the invisible keypoints by masking the loss.
This results in a regression-based one-stage multi-person human pose estimator similar to the slow-RCNN version counterparts Toshev \etal~\cite{toshev2014deeppose} and Sun \etal~\cite{sun2017compositional}.

To refine the keypoints, we further estimate $k$ human joint heatmaps $\hat \Phi \in \mathcal{R}^{\frac{W}{R} \times \frac{H}{R} \times k}$ using standard bottom-up multi-human pose estimation~\cite{cao2017realtime,newell2017associative,papandreou2018personlab}.
We train the human joint heatmap with focal loss and local pixel offset analogous to the center detection discussed in Section.~\ref{sec:prelim}.

We then snap our initial predictions to the closest detected keypoint on this heatmap.
Here, our center offset acts as a grouping cue, to assign individual keypoint detections to their closest person instance. 
Specifically, let $(\hat x,\hat y)$ be a detected center point.
We first regress to all joint locations $l_j = (\hat x,\hat y) + \hat J_{\hat x\hat yj}$ for $j \in 1 \ldots k$.
We also extract all keypoint locations $L_j = \{\tilde l_{ji}\}_{i=1}^{n_j}$ with a confidence $>0.1$ for each joint type $j$ from the corresponding heatmap $\hat \Phi_{\cdot\cdot j}$.
We then assign each regressed location $l_j$ to its closest detected keypoint ${\arg\min}_{l \in L_j} (l - l_j)^2$ considering only joint detections within the bounding box of the detected object.

\section{Implementation details}

We experiment with 4 architectures: ResNet-18, ResNet-101~\cite{xiao2018simple}, DLA-34~\cite{yu2018deep}, and Hourglass-104~\cite{Law_2018_ECCV}.
We modify both ResNets and DLA-34 using deformable convolution layers~\cite{Dai_2017_ICCV} and use the Hourglass network as is.

\label{sec:networks}
\paragraph{Hourglass}
The stacked Hourglass Network~\cite{newell2016stacked,Law_2018_ECCV} downsamples the input by $4\times$,  followed by two sequential hourglass modules.
Each hourglass module is a symmetric 5-layer down- and up-convolutional network with skip connections.
This network is quite large, but generally yields the best keypoint estimation performance.

\paragraph{ResNet}
Xiao et al.~\cite{xiao2018simple} augment a standard residual network~\cite{he2016deep} with three up-convolutional networks to allow for a higher-resolution output (output stride $4$).
We first change the channels of the three upsampling layers to $256, 128, 64$, respectively, to save computation.
We then add one $3\times3$ deformable convolutional layer before each up-convolution with channel $256, 128, 64$, respectively.
The up-convolutional kernels are initialized as bilinear interpolation.
See supplement for a detailed architecture diagram.

\paragraph{DLA}
Deep Layer Aggregation (DLA)~\cite{yu2018deep} is an image classification network with hierarchical skip connections.
We utilize the fully convolutional upsampling version of DLA for dense prediction, 
which uses iterative deep aggregation to increase feature map resolution symmetrically. 
We augment the skip connections with deformable convolution~\cite{zhu2018deformable} from lower layers to the output.
Specifically, we replace the original convolution with $3\times3$ deformable convolution at every upsampling layer.
See supplement for a detailed architecture diagram.

We add one $3\times3$ convolutional layer with $256$ channel before each output head.
A final $1 \times 1$ convolution then produces the desired output.
We provide more details in the supplementary material.

\paragraph{Training}
We train on an input resolution of $512 \times 512$.
This yields an output resolution of $128 \times 128$ for all the models.
We use random flip, random scaling (between 0.6 to 1.3), cropping, and color jittering as data augmentation, and use Adam\cite{kingma2014adam} to optimize the overall objective.
We use no augmentation to train the 3D estimation branch, as cropping or scaling changes the 3D measurements.
For the residual networks and DLA-34, we train with a batch-size of 128 (on 8 GPUs) and learning rate 5e-4 for 140 epochs, with learning rate dropped $10 \times$ at 90 and 120 epochs, respectively (following ~\cite{xiao2018simple}).
For Hourglass-104, we follow ExtremeNet~\cite{zhou2019bottomup} and use batch-size 29 (on 5 GPUs, with master GPU batch-size 4) and learning rate 2.5e-4 for 50 epochs with $10 \times$ learning rate dropped at the 40 epoch.
For detection, we fine-tune the Hourglass-104 from ExtremeNet~\cite{zhou2019bottomup} to save computation.
The down-sampling layers of Resnet-101 and DLA-34 are initialized with ImageNet pretrain and the up-sampling layers are randomly initialized.
Resnet-101 and DLA-34 train in 2.5 days on 8 TITAN-V GPUs, while Hourglass-104 requires 5 days.

\paragraph{Inference}
We use three levels of test augmentations: no augmentation, flip augmentation, and flip and multi-scale (0.5, 0.75, 1, 1.25, 1.5).
For flip, we average the network outputs before decoding bounding boxes. For multi-scale, we use NMS to merge results.
These augmentations yield different speed-accuracy trade-off, as is shown in the next section.

{
\setlength{\tabcolsep}{5pt}
\begin{table*}[t]
\centering
\begin{tabular}{l@{}c@{\ \ }ccccccc}
\toprule
& Backbone & FPS & $AP$ & $AP_{50}$ & $AP_{75}$ & $AP_{S}$ & $AP_{M}$ & $AP_{L}$  \\
\midrule
        MaskRCNN~\cite{he2017mask} & ResNeXt-101 & \textbf{11} & 39.8 & 62.3 & 43.4 & 22.1 & 43.2 & 51.2 \\
        Deform-v2~\cite{zhu2018deformable} & ResNet-101 & - & 46.0 & 67.9 & 50.8 & 27.8 & 49.1 & 59.5 \\
        SNIPER~\cite{sniper2018} & DPN-98 & \textit{2.5} & 46.1 & 67.0 & 51.6 & 29.6 & 48.9 & 58.1  \\
        PANet~\cite{liu2018path} & ResNeXt-101 & - &  47.4 & 67.2 & 51.8 & 30.1& \textbf{51.7}& 60.0\\ 
        TridentNet~\cite{li2019scale} & ResNet-101-DCN & 0.7 &  \textbf{48.4} & \textbf{69.7} & \textbf{53.5} & \textbf{31.8} & 51.3 & \textbf{60.3}\\
        \midrule
        YOLOv3~\cite{redmon2018yolov3} & DarkNet-53 & 20 &  33.0 & 57.9 & 34.4&  18.3 & 25.4 & 41.9 \\
        RetinaNet~\cite{lin2018focal} & ResNeXt-101-FPN & 5.4 & 40.8 & 61.1 & 44.1 &  24.1 & 44.2 & 51.2 \\
        RefineDet~\cite{zhang2018single} & ResNet-101 & - &  36.4\ /\ 41.8 & 57.5\ /\ 62.9 & 39.5\ /\ 45.7 &  16.6\ /\ 25.6 & 39.9\ /\ 45.1 & 51.4\ /\ 54.1\\
        CornerNet~\cite{Law_2018_ECCV} & Hourglass-104 & 4.1 &  40.5\ /\ 42.1 & 56.5\ /\ 57.8 & 43.1\ /\ 45.3 &  19.4\ /\ 20.8 & 42.7\ /\ 44.8 & \textbf{53.9}\ /\ 56.7 \\
        ExtremeNet~\cite{zhou2019bottomup} & Hourglass-104 & 3.1 &  40.2\ /\ 43.7 & 55.5\ /\ 60.5 & 43.2\ /\ 47.0 &  20.4\ /\ 24.1 & 43.2\ /\ 46.9 & 53.1\ /\ 57.6 \\
        FSAF~\cite{zhu2019feature} & ResNeXt-101 & \textit{2.7} & \textbf{42.9}\ /\ 44.6 & \textbf{63.8}\ /\ \textbf{65.2} & \textbf{46.3}\ /\ 48.6 & \textbf{26.6}\ /\ \textbf{29.7} & \textbf{46.2}\ /\ \textbf{47.1} & 52.7\ /\ 54.6 \\
        CenterNet-DLA & DLA-34 &  \textbf{28} &  39.2\ /\ 41.6 & 57.1\ /\ 60.3 & 42.8\ /\ 45.1 &  19.9\ /\ 21.5 & 43.0\ /\ 43.9 & 51.4\ /\ 56.0 \\
        CenterNet-HG & Hourglass-104 & 7.8 & 42.1\ /\ \textbf{45.1} & 61.1\ /\ 63.9 & 45.9\ /\ \textbf{49.3} & 24.1\ /\ 26.6 & 45.5\ /\ \textbf{47.1} & 52.8\ /\ \textbf{57.7}\\
\bottomrule
\end{tabular}
\caption{State-of-the-art comparison on COCO test-dev. Top: two-stage detectors; bottom: one-stage detectors. We show single-scale / multi-scale testing for most one-stage detectors. Frame-per-second (FPS) were measured on the same machine whenever possible. Italic FPS highlight the cases, where the performance measure was copied from the original publication. A dash indicates methods for which neither code and models, nor public timings were available.}
\label{table:main}
\vspace{-1em}
\end{table*}
\setlength{\tabcolsep}{1.4pt}
}

\section{Experiments}

We evaluate our object detection performance on the MS COCO dataset~\cite{lin2014microsoft}, which contains 118k training images (train2017), 5k validation images (val2017) and 20k hold-out testing images (test-dev).
We report average precision over all IOU thresholds (AP), AP at IOU thresholds 0.5($AP_{50}$) and 0.75 ($AP_{75}$). 
The supplement contains additional experiments on PascalVOC~\cite{pascal-voc-2012}.

\subsection{Object detection}
Table~\ref{tab:network} shows our results on COCO validation with different backbones and testing options, while Figure~\ref{fig:tradeoff} compares CenterNet with other real-time detectors.
The running time is tested on our local machine, with Intel Core i7-8086K CPU, Titan Xp GPU, Pytorch 0.4.1, CUDA 9.0, and CUDNN 7.1.
We download code and pre-trained models\footnote{\url{https://github.com/facebookresearch/Detectron}}\footnote{\url{https://github.com/pjreddie/darknet}} to test run time for each model on the same machine.

Hourglass-104 achieves the best accuracy at a relatively good speed, with a $42.2\%$ AP in $7.8$ FPS.
On this backbone, CenterNet outperforms CornerNet~\cite{Law_2018_ECCV} ($40.6\%$ AP in $4.1$ FPS) and ExtremeNet~\cite{zhou2019bottomup}($40.3\%$ AP in $3.1$ FPS) in both speed and accuracy.
The run time improvement comes from fewer output heads and a simpler box decoding scheme.
Better accuracy indicates that center points are easier to detect than corners or extreme points.

Using ResNet-101, we outperform RetinaNet~\cite{lin2018focal} with the same network backbone.
We only use deformable convolutions in the upsampling layers, which does not affect RetinaNet.
We are more than twice as fast at the same accuracy (CenterNet $34.8\%$AP in $45$ FPS (input $512\times512$) vs. RetinaNet $34.4\%$AP in $18$ FPS (input $500\times800$)).
Our fastest ResNet-18 model also achieves a respectable performance of $28.1\%$ COCO AP at $142$ FPS.

DLA-34 gives the best speed/accuracy trade-off. 
It runs at $52 $FPS with $37.4\%$AP.
This is more than twice as fast as YOLOv3~\cite{redmon2018yolov3} and $4.4\%$AP more accurate.
With flip testing, our model is still faster than YOLOv3~\cite{redmon2018yolov3} and achieves accuracy levels of Faster-RCNN-FPN~\cite{ren2015faster} (CenterNet $39.2\%$ AP  in $28$ FPS vs Faster-RCNN $39.8\%$ AP in $11$ FPS).

\paragraph{State-of-the-art comparison}
We compare with other state-of-the-art detectors in COCO test-dev in Table~\ref{table:main}.
With multi-scale evaluation, CenterNet with Hourglass-104 achieves an AP of $45.1\%$, outperforming all existing one-stage detectors.
Sophisticated two-stage detectors~\cite{zhu2018deformable,sniper2018,liu2018path,li2019scale} are more accurate, but also slower.
There is no significant difference between CenterNet and sliding window detectors for different object sizes or IoU thresholds.
CenterNet behaves like a regular detector, just faster.

\begin{table*}[t]
\setlength{\tabcolsep}{1.4pt}
\begin{subfigure}[t]{0.28\linewidth}
\centering
\begin{tabular}{l c c c c}
\hline
Resolution & AP & $AP_{50}$ & $AP_{75}$ & Time \\
\hline
Original & \textbf{36.3} & 54.0 & \textbf{39.6} & 19 \\
$512$& 36.2 & \textbf{54.3} & 38.7 & 16 \\
$384$ & 33.2 & 50.5 & 35.0 & \textbf{11} \\
\hline
\end{tabular}
\caption{Testing resolution: Lager resolutions perform better but run slower.} \label{tab:design:resolution}
\end{subfigure}
\hspace{2mm}
\begin{subfigure}[t]{0.23\linewidth}
\centering
\begin{tabular}{l c c c}
\hline
$\lambda_{size}$ & AP & $AP_{50}$ & $AP_{75}$ \\
\hline
$0.2$ & 33.5 & 49.9 & 36.2\\
$0.1$ & \textbf{36.3} & 54.0 & \textbf{39.6} \\
$0.02$ & 35.4 & \textbf{54.6} & 37.9 \\
\hline
\end{tabular}
\caption{Size regression weight. $\!\lambda_{size}\!\le\!0.1$ yields good results.}
\label{tab:design:lossweight}
\end{subfigure}
\hspace{2mm}
\begin{subfigure}[t]{0.23\linewidth}
\centering
\begin{tabular}{l c c c}
\hline
Loss & AP & $AP_{50}$ & $AP_{75}$ \\
\hline
l1 & \textbf{36.3} & \textbf{54.0} & \textbf{39.6} \\
smooth l1 & 33.9 & 50.9 & 36.8 \\
\hline
\multicolumn{4}{c}{~}\\
\end{tabular}
\caption{Regression loss. L1 loss works better than Smooth L1.}
\label{tab:design:regloss}
\end{subfigure}
\hspace{2mm}
\begin{subfigure}[t]{0.2\linewidth}
\centering
\begin{tabular}{l c c c}
\hline
Epoch & AP & $AP_{50}$ & $AP_{75}$ \\
\hline
140& 36.3 & 54.0 & 39.6 \\
230& \textbf{37.4} & \textbf{55.1} & \textbf{40.8} \\
\hline
\multicolumn{4}{c}{~}\\
\end{tabular}
\caption{Training schedule. Longer performs better.}
\label{tab:design:schedule}
\end{subfigure}
\vspace{-0.5em}
\caption{Ablation of design choices on COCO validation set. The results are shown in COCO AP, time in milliseconds.}
\label{tab:design}
\vspace{-1.5em}
\setlength{\tabcolsep}{2pt}
\end{table*}

\begin{table*}[!hb]
\centering
\begin{tabular}{l c@{\ \ } c@{\ \ } c@{\ \ } c@{\ \ } c@{\ \ } c@{\ \ } c@{\ \ } c@{\ \ } c}
\hline
& \multicolumn{3}{c}{AP} & \multicolumn{3}{c}{AOS} & \multicolumn{3}{c}{BEV AP} \\
 & Easy & Mode & Hard & Easy & Mode & Hard & Easy & Mode & Hard \\
\hline
Deep3DBox~\cite{mousavian20173d} & 98.8 & 97.2 & 81.2 & 98.6 & 96.7 & 80.5 & 30.0 & 23.7 & 18.8 \\
Ours & 90.2$\pm$1.2 & 80.4$\pm$1.4 & 71.1$\pm$1.6 & 85.3$\pm$1.7 & 75.0$\pm$1.6 & 66.2$\pm$1.8 & 31.4$\pm$3.7 & 26.5$\pm$1.6 & 23.8$\pm$2.9 \\
\hline
Mono3D~\cite{chen2016monocular} & 95.8 & 90.0 & 80.6 & 93.7 & 87.6 & 78.0 & 30.5 & 22.4 & 19.1 \\
Ours & 97.1$\pm$0.3 & 87.9$\pm$0.1 & 79.3$\pm$0.1 & 93.4$\pm$0.7 & 83.9$\pm$0.5 & 75.3$\pm$0.4 & 31.5$\pm$2.0 & 29.7$\pm$0.7 & 28.1$\pm$4.6 \\
\hline
\end{tabular}
\vspace{-0.5em}
\caption{KITTI evaluation. We show 2D bounding box AP, average orientation score (AOS), and bird eye view (BEV) AP  on different validation splits. Higher is better.}
\label{tab:kittival}
\vspace{-0.5em}
\end{table*}

\subsubsection{Additional experiments}

In unlucky circumstances, two different objects might share the same center, if they perfectly align.
In this scenario, CenterNet would only detect one of them.
We start by studying how often this happens in practice and put it in relation to missing detections of competing methods.

\paragraph{Center point collision}
In the COCO training set, there are $614$ pairs of objects that collide onto the same center point at stride $4$.
There are $860001$ objects in total, hence CenterNet is unable to predict $<0.1\%$ of objects due to collisions in center points.
This is much less than slow- or fast-RCNN miss due to imperfect region proposals~\cite{uijlings2013selective} ($\sim 2\%$), and fewer than anchor-based methods miss due to insufficient anchor placement~\cite{ren2015faster} ($20.0\%$ for Faster-RCNN with $15$ anchors at $0.5$ IOU threshold).
In addition, $715$ pairs of objects have bounding box IoU $>0.7$ and would be assigned to two anchors, hence a center-based assignment causes fewer collisions.

\paragraph{NMS}
To verify that IoU based NMS is not needed for CenterNet, 
we ran it as a post-processing step on our predictions.
For DLA-34 (flip-test), the AP improves from $39.2\%$ to $39.7\%$. 
For Hourglass-104, the AP stays at $42.2\%$.
Given the minor impact, we do not use it.
\\ \\
Next, we ablate the new hyperparameters of our model.
All the experiments are done on DLA-34.

\paragraph{Training and Testing resolution}
During training, we fix the input resolution to $512 \times 512$.
During testing, we follow CornerNet~\cite{Law_2018_ECCV} to keep the original image resolution and zero-pad the input to the maximum stride of the network.
For ResNet and DLA, we pad the image with up to 32 pixels, for HourglassNet, we use 128 pixels.
As is shown in Table.~\ref{tab:design:resolution}, keeping the original resolution is slightly better than fixing test resolution.
Training and testing in a lower resolution ($384 \times 384$) runs $1.7$ times faster but drops $3$AP.

\paragraph{Regression loss}
We compare a vanilla L1 loss to a Smooth L1~\cite{girshick2015fast} for size regression.
Our experiments in Table~\ref{tab:design:regloss} show that L1 is considerably better than Smooth L1.
It yields a better accuracy at fine-scale, which the COCO evaluation metric is sensitive to.
This is independently observed in keypoint regression~\cite{sun2017compositional,sun2018integral}.

\paragraph{Bounding box size weight}
We analyze the sensitivity of our approach to the loss weight $\lambda_{size}$.
Table~\ref{tab:design:lossweight} shows $0.1$ gives a good result.
For larger values, the AP degrades significantly, due to the scale of the loss ranging from $0$ to output size $w / R$ or $h / R$, instead of $0$ to $1$.
However, the value does not degrade significantly for lower weights.

\paragraph{Training schedule}
By default, we train the keypoint estimation network for $140$ epochs with a learning rate drop at $90$ epochs.
If we double the training epochs before dropping the learning rate, the performance further increases by $1.1$ AP (Table~\ref{tab:design:schedule}), at the cost of a much longer training schedule.
To save computational resources (and polar bears), we use $140$ epochs in ablation experiments, but stick with $230$ epochs for DLA when comparing to other methods.

Finally, we tried a multiple ``anchor'' version of CenterNet by regressing to more than one object size.
The experiments did not yield any success. See supplement.

\subsection{3D detection}

We perform 3D bounding box estimation experiments on KITTI dataset~\cite{Geiger2012CVPR}, which contains carefully annotated 3D bounding box for vehicles in a driving scenario.
KITTI contains $7841$ training images and we follow standard training and validation splits in literature~\cite{xiang2017subcategory,chen20153d}.
The evaluation metric is the average precision for cars at $11$ recalls ($0.0$ to $1.0$ with $0.1$ increment) at IOU threshold $0.5$, as in object detection~\cite{pascal-voc-2012}. 
We evaluate IOUs based on 2D bounding box (AP), orientation (AOP), and Bird-eye-view bounding box (BEV AP).
We keep the original image resolution and pad to $1280 \times 384$ for both training and testing.
The training converges in $70$ epochs, with learning rate dropped at the $45$ and $60$ epoch, respectively.
We use the DLA-34 backbone and set the loss weight for depth, orientation, and dimension to $1$.
All other hyper-parameters are the same as the detection experiments.

Since the number of recall thresholds is quite small, the validation AP fluctuates by up to $10\%$ AP.
We thus train $5$ models and report the average with standard deviation.

We compare with slow-RCNN based Deep3DBox~\cite{mousavian20173d} and Faster-RCNN based method Mono3D~\cite{chen2016monocular}, on their specific validation split.
As is shown in Table~\ref{tab:kittival}, our method performs on-par with its counterparts in AP and AOS and does slightly better in BEV.
Our CenterNet is two orders of magnitude faster than both methods.

\subsection{Pose estimation}

Finally, we evaluate CenterNet on human pose estimation in the MS COCO dataset~\cite{lin2014microsoft}.
We evaluate keypoint AP, which is similar to bounding box AP but replaces the bounding box IoU with object keypoint similarity.
We test and compare with other methods on COCO test-dev.

We experiment with DLA-34 and Hourglass-104, both fine-tuned from center point detection.
DLA-34 converges in 320 epochs (about 3 days on 8GPUs) and Hourglass-104 converges in 150 epochs (8 days on 5 GPUs).
All additional loss weights are set to $1$.
All other hyper-parameters are the same as object detection.

The results are shown in Table~\ref{table:human}.
Direct regression to keypoints performs reasonably, but not at state-of-the-art.
It struggles particularly in high IoU regimes.
Projecting our output to the closest joint detection improves the results throughout, and performs competitively with state-of-the-art multi-person pose estimators~\cite{cao2017realtime,newell2017associative,he2017mask,papandreou2018personlab}.
This verifies that CenterNet is general, easy to adapt to a new task.

\begin{figure*}[t]
\centering
     \begin{tabular}{cccc}
     \includegraphics[trim={0 0 0 0.3cm}, clip, width=0.24\textwidth]{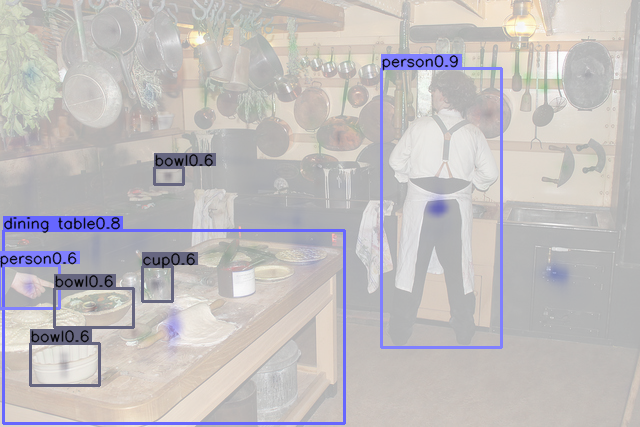}
      &\includegraphics[trim={0 0 0 0}, clip, width=0.24\textwidth]{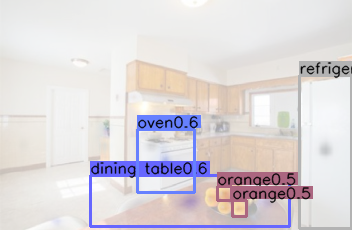}
      &\includegraphics[trim={0 0 0 2.2cm}, clip, width=0.24\textwidth]{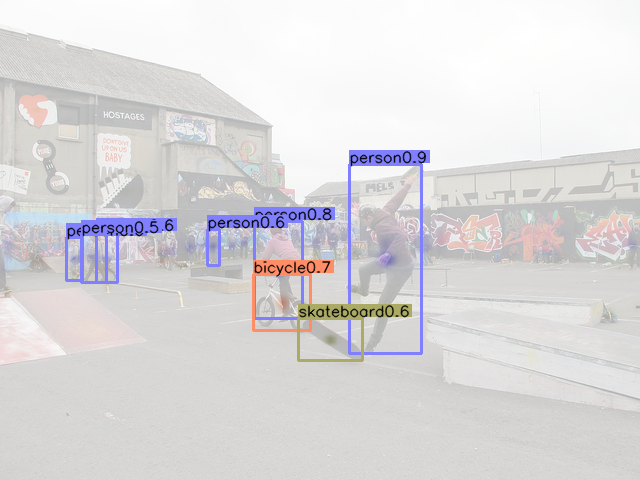}
      & \includegraphics[trim={0 0 0 2.3cm}, clip, width=0.24\textwidth]{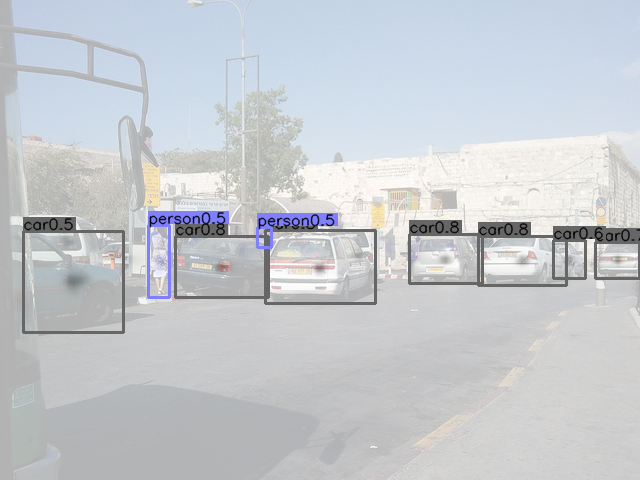} \\
     \includegraphics[width=0.24\textwidth]{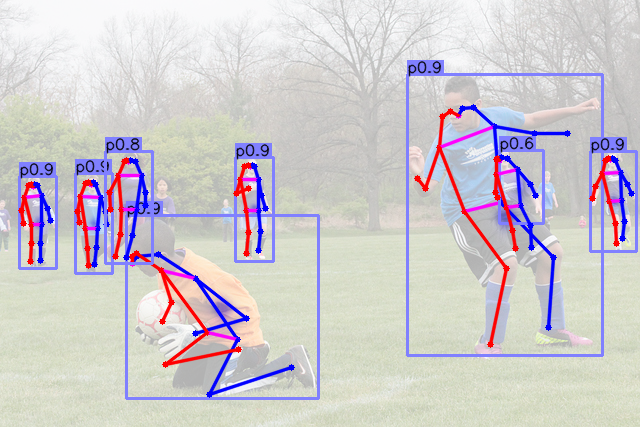}
      &\includegraphics[width=0.24\textwidth]{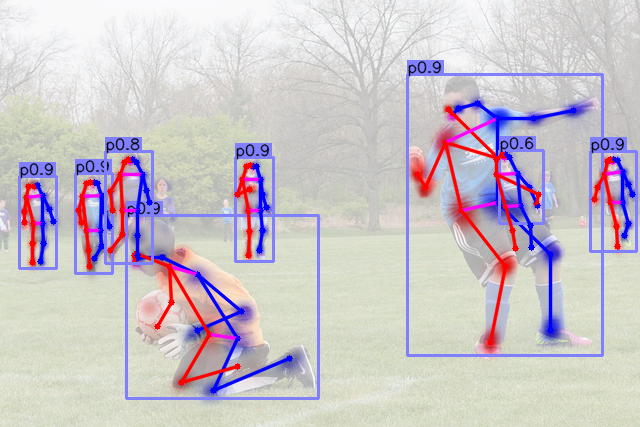}
      &\includegraphics[width=0.24\textwidth]{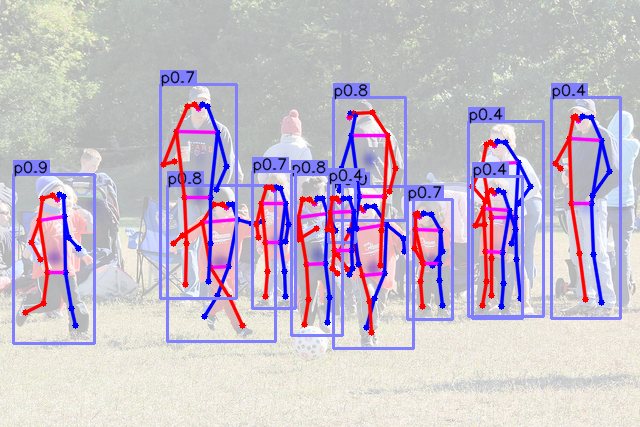}
      &\includegraphics[width=0.24\textwidth]{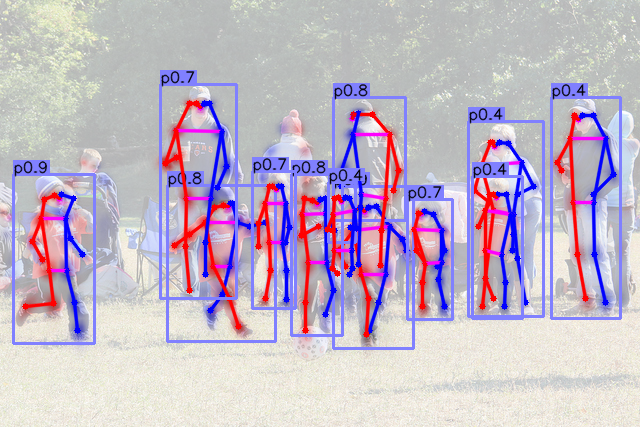} \\
     \includegraphics[width=0.24\textwidth]{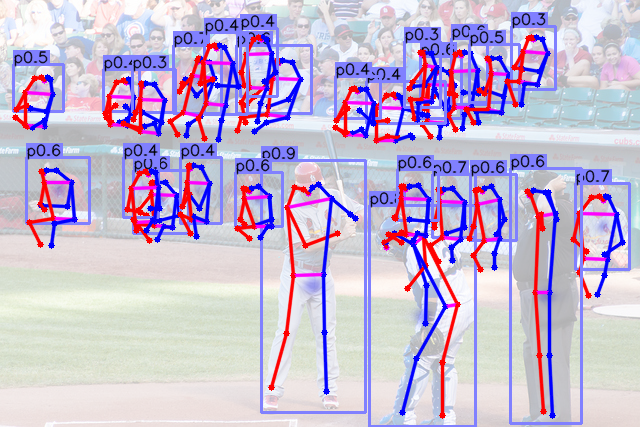}
      &\includegraphics[width=0.24\textwidth]{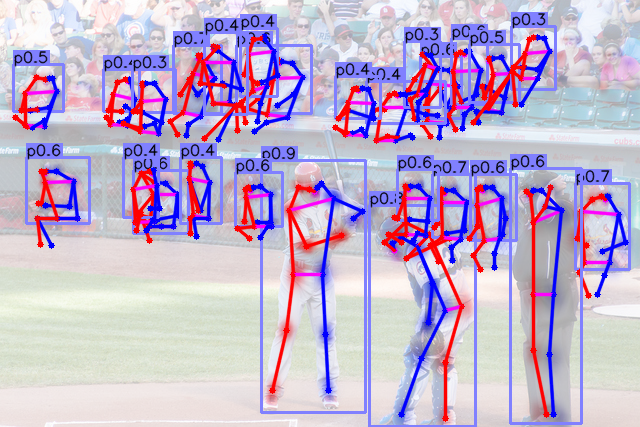}
      &\includegraphics[trim={0 0 0 1.9cm}, clip, width=0.24\textwidth]{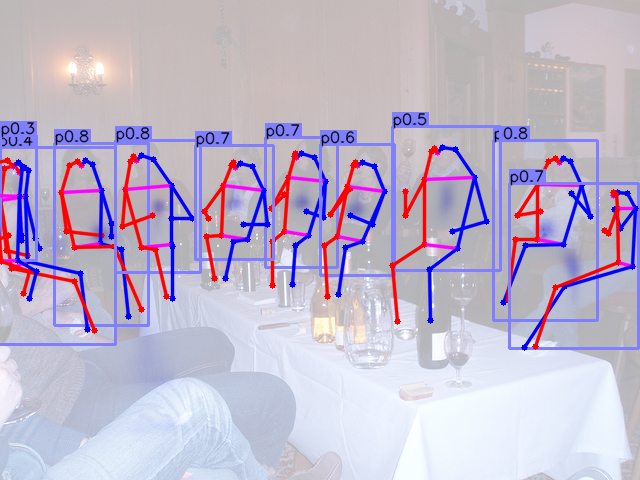}
      &\includegraphics[trim={0 0 0 1.9cm}, clip, width=0.24\textwidth]{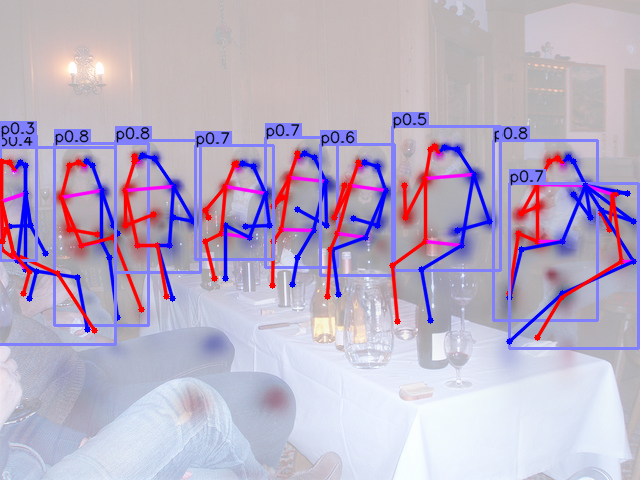} \\
      \multicolumn{2}{c}{\includegraphics[width=0.49\textwidth]{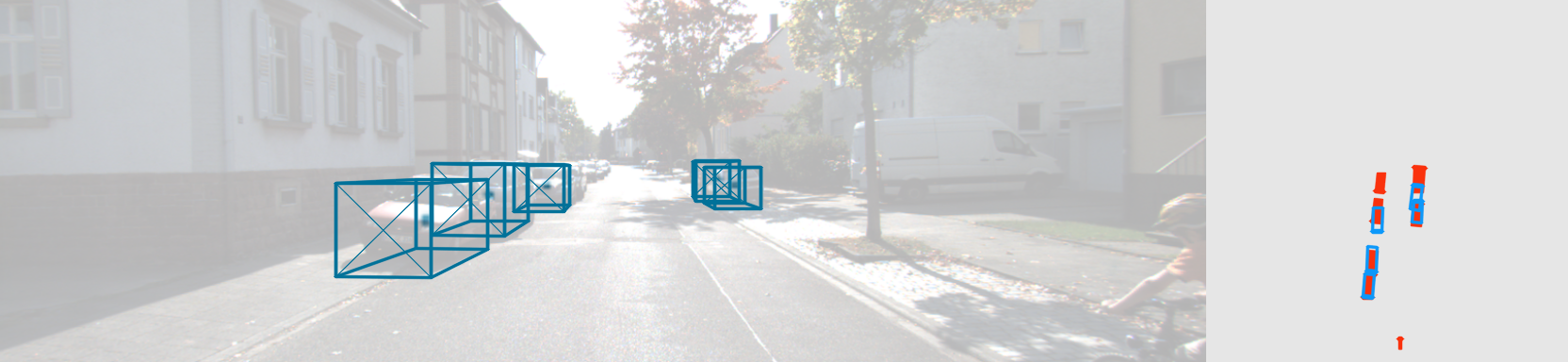}}
      &\multicolumn{2}{c}{\includegraphics[width=0.49\textwidth]{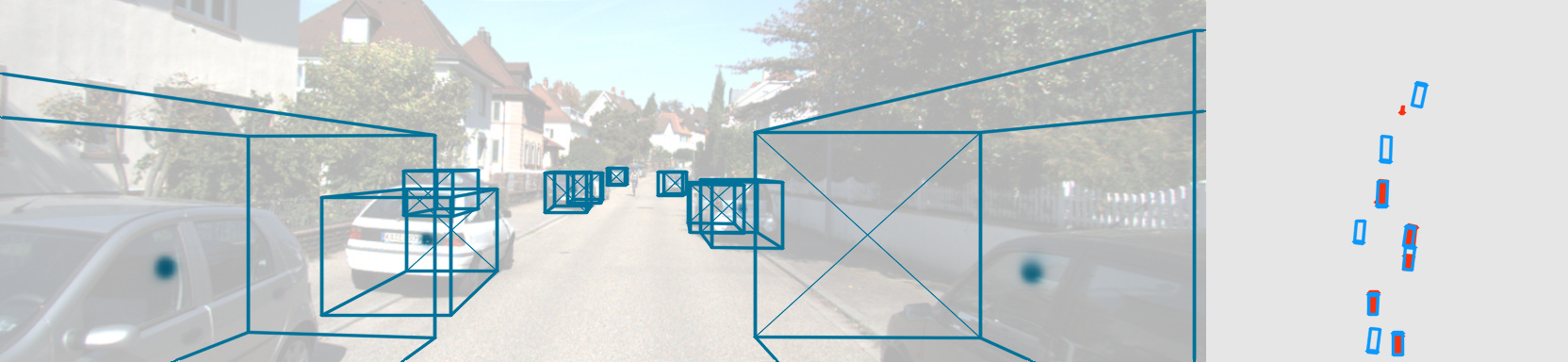}} \\
      \multicolumn{2}{c}{\includegraphics[width=0.49\textwidth]{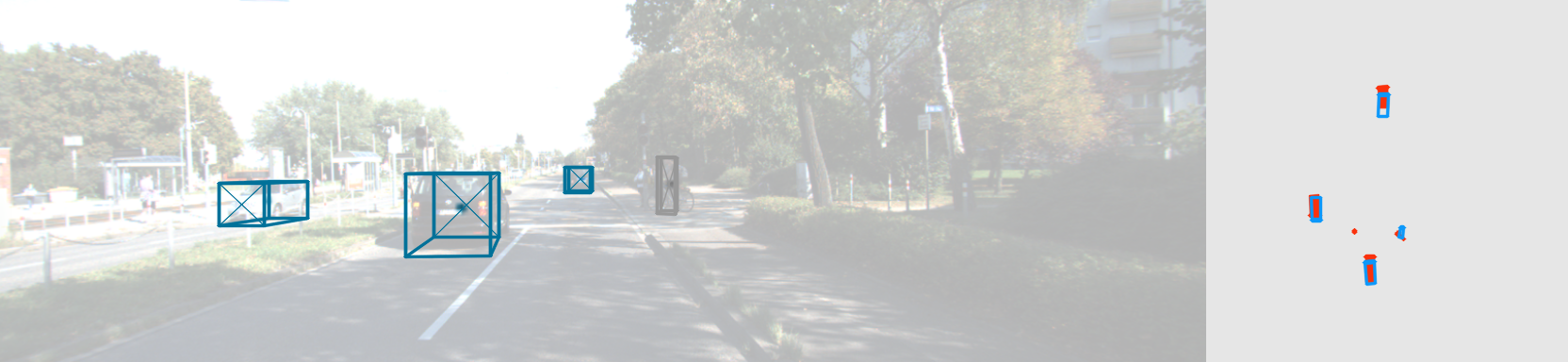}}
      &\multicolumn{2}{c}{\includegraphics[width=0.49\textwidth]{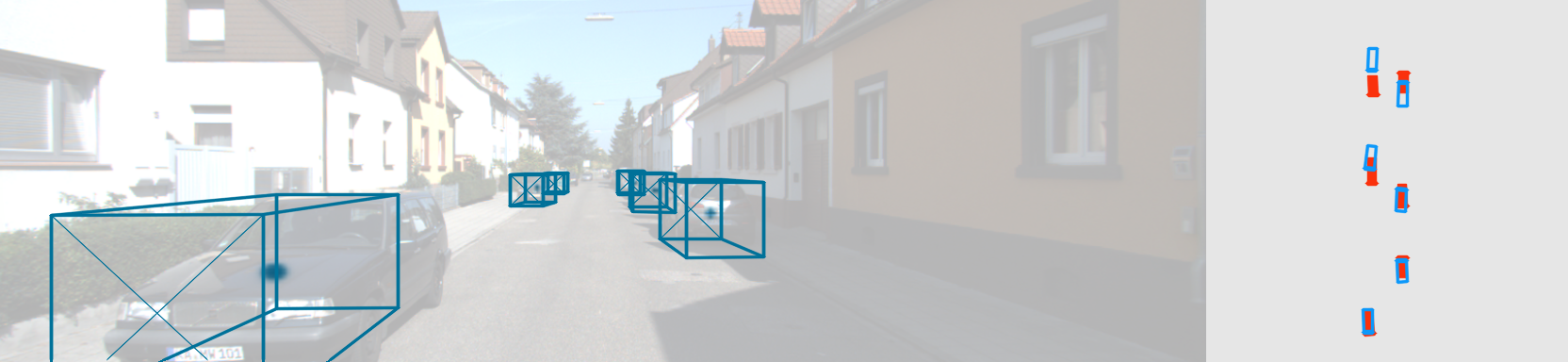}} \\
      \end{tabular}
      \caption{Qualitative results. All images were picked thematically without considering our algorithms performance. \emph{First row:} object detection on COCO validation. \emph{Second and third row:} Human pose estimation on COCO validation. For each pair, we show the results of center offset regression (left) and heatmap matching (right). \emph{fourth and fifth row:} 3D bounding box estimation on KITTI validation. We show projected bounding box (left) and bird eye view map (right). The ground truth detections are shown in solid red solid box. The center heatmap and 3D boxes are shown overlaid on the original image.}
      \label{fig:demo}
\end{figure*}

Figure~\ref{fig:demo} shows qualitative examples on all tasks.

\setlength{\tabcolsep}{2pt}
\begin{table}[t]
\centering
\begin{tabular}{l@{\ \ }c@{\ \ }c@{\ \ }c@{\ \ }c@{\ \ }c}
\hline
& $AP^{kp}$ & $AP^{kp}_{50}$ & $AP^{kp}_{75}$ & $AP^{kp}_{M}$ & $AP^{kp}_{L}$ \\
\hline
CMU-Pose~\cite{cao2017realtime} & 61.8 & 84.9 & 67.5 & 58.0 & 70.4 \\
Pose-AE~\cite{newell2017associative} & 62.8 & 84.6 & 69.2 & 57.5 & 70.6 \\
Mask-RCNN~\cite{he2017mask} & 63.1 & 87.3 & 68.7 & 57.8 & 71.4 \\
PersonLab~\cite{papandreou2018personlab} & 66.5 & 85.5 & 71.3 & 62.3 & 70.0 \\
\hline
DLA-reg & 51.7 & 81.4 & 55.2 & 44.6 & 63.0 \\
HG-reg & 55.0 & 83.5 & 59.7 & 49.4 & 64.0 \\
DLA-jd & 57.9 & 84.7 & 63.1 & 52.5 & 67.4 \\
HG-jd & 63.0 & 86.8 & 69.6 & 58.9 & 70.4 \\
\hline
\end{tabular}
\caption{Keypoint detection on COCO test-dev. -reg/ -jd are for direct center-out offset regression and matching regression to the closest joint detection, respectively. The results are shown in COCO keypoint AP. Higher is better.}
\label{table:human}
\vspace{-1em}
\end{table}
\setlength{\tabcolsep}{1.4pt}

\section{Conclusion}
In summary, we present a new representation for objects: as points.
Our CenterNet object detector builds on successful keypoint estimation networks, finds object centers, and regresses to their size.
The algorithm is simple, fast, accurate, and end-to-end differentiable without any NMS post-processing.
The idea is general and has broad applications beyond simple two-dimensional detection.
CenterNet can estimate a range of additional object properties, such as pose, 3D orientation, depth and extent, in one single forward pass.
Our initial experiments are encouraging and open up a new direction for real-time object recognition and related tasks.
\newpage
\begin{spacing}{1.1}
{
\bibliographystyle{ieee}
\bibliography{ref}
}
\end{spacing}
\newpage
\appendix

\twocolumn[{
\maketitle
\centering
   \includegraphics[width=0.95\linewidth]{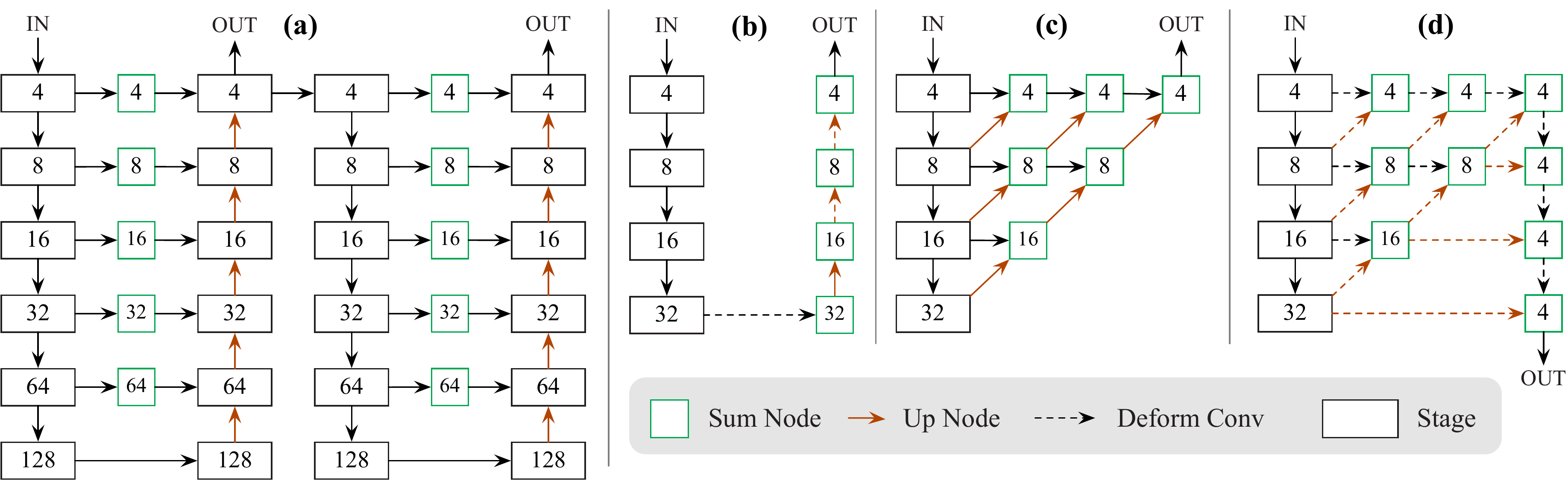}
   \captionof{figure}{Model diagrams. The numbers in the boxes represent the stride to the image. (a): Hourglass Network~\cite{Law_2018_ECCV}. We use it as is in CornerNet~\cite{Law_2018_ECCV}. (b): ResNet with transpose convolutions~\cite{xiao2018simple}. We add one $3\times3$ deformable convolutional layer~\cite{zhu2018deformable} before each up-sampling layer. Specifically, we first use deformable convolution to change the channels and then use transposed convolution to upsample the feature map (such two steps are shown separately in $32\rightarrow16$. We show these two steps together as a dashed arrow for $16\rightarrow8$ and $8\rightarrow4$). (c): The original DLA-34~\cite{yu2018deep} for semantic segmentation. (d): Our modified DLA-34. We add more skip connections from the bottom layers and upgrade every convolutional layer in upsampling stages to deformable convolutional layer.}
\label{fig:arch}
\vspace{5mm}
}]
\section*{Appendix A: Model Architecture}
See figure.~\ref{fig:arch} for diagrams of the architectures.

\section*{Appendix B: 3D BBox Estimation Details}

Our network outputs maps for depths $\hat{D} \in R^{\frac{W}{R} \times \frac{H}{R}}$, 3d dimensions $\hat{\Gamma} \in R^{\frac{W}{R} \times \frac{H}{R} \times 3}$, and orientation encoding $\hat{A} \in R^{\frac{W}{R} \times \frac{H}{R} \times 8}$. 
For each object instance $k$, we extract the output values from the three output maps at the ground truth center point location:
$\hat{d}_k \in R$, $\hat{\gamma}_k \in R^3$, $\hat{\alpha}_k \in R^8$.
The depth is trained with L1 loss after converting the output to the absolute depth domain:
\begin{equation}
    L_{dep} = \frac{1}{N}\sum_{k = 1}^N{|\frac{1}{\sigma(\hat{d}_k)} - 1 - d_k|}
\end{equation}
where $d_k$ is the groud truth absolute depth (in meter).
Similarly, the 3D dimension is trained with L1 Loss in absolute metric:
\begin{equation}
    L_{dim} = \frac{1}{N}\sum_{k = 1}^N{|\hat{\gamma}_k - \gamma_k|}
\end{equation}
where $\gamma_k$ is the object height, width, and length in meter.

The orientation $\theta$ is a single scalar by default.
Following Mousavian \etal~\cite{mousavian20173d,hu2018joint}, We use an $8$-scalar encoding to ease learning.
The $8$ scalars are divided into two groups, each for an angular bin.
One bin is for angles in $B_1 = [-\frac{7\pi}{6}, \frac{\pi}{6}]$ and the other is for angles in $B_2 = [-\frac{\pi}{6}, \frac{7\pi}{6}]$. 
Thus we have $4$ scalars for each bin.
Within each bin, $2$ of the scalars $b_i \in R^2$ are used for softmax classification (if the orientation falls into to this bin $i$).
And the rest $2$ scalars $a_i \in R^2$ are for the $\sin$ and $\cos$ value of in-bin offset (to the bin center $m_i$).
I.e., $\hat{\alpha} = [\hat{b}_1, \hat{a}_1, \hat{b}_2, \hat{a}_2]$
The classification are trained with softmax and the angular values are trained with L1 loss:
\begin{equation}
    L_{ori} = \frac{1}{N}\sum_{k = 1}^N{\sum_{i = 1}^2  {(softmax(\hat{b}_i, c_i) + c_i|\hat{a}_i - a_i}|)}
\end{equation}
where $c_i = \mathbbm{1}(\theta \in B_i)$, $a_i = (\sin{(\theta - m_i)}, \cos{(\theta - m_i)})$.
$\mathbbm{1}$ is the indicator function.
The predicted orientation $\theta$ is decoded from the $8$-scalar encoding by
\begin{equation}
    \hat{\theta} = arctan2(\hat{a}_{j1}, \hat{a}_{j2}) + m_j
\end{equation}
where $j$ is the bin index which has a larger classification score.

\section*{Appendix C: Collision Experiment Details}
We analysis the annotations of COCO training set to show how often the collision cases happen.
COCO training set (train 2017) contains $N=118287$ images and $M=860001$ objects (with $M_S = 356340$ small objects, $M_M = 295163$ medium objects, and $M_L = 208498$ large objects) in $C=80$ categories.
Let the $i$-th bounding box of image $k$ of category $c$ be  $bb^{(kci)} = (x_1^{(kci)}, y_1^{(kci)}, x_2^{(kci)}, y_2^{(kci)})$, 
its center after the $4 \times$ stride is $p^{kci} = (\lfloor \frac{1}{4} \cdot \frac{x_1^{(kci)} + x_2^{(kci)}}{2}\rfloor, \lfloor \frac{1}{4} \cdot \frac{y_1^{(kci)} + y_2^{(kci)}}{2}\rfloor)$.
And Let $n^{(kc)}$ be the number of object of category $c$ in image $k$.
The number of center point collisions is calculated by:
\begin{equation}
    N_{center} = \sum_{k=1}^{N} \sum_{c=1}^{C} \sum_{i=1}^{n^{(kc)}} \sum_{j=i+1}^{n^{(kc)}} \mathbbm{1} (p^{kci} = p^{kcj})
\end{equation}
We get $N_{center} = 614$ on the dataset.

Similarly, we calculate the IoU based collision by
\begin{equation}
    N_{IoU@t} = \sum_{k=1}^{N} \sum_{c=1}^{C} \sum_{i=1}^{n^{(kc)}} \sum_{j=i+1}^{n^{(kc)}} \mathbbm{1} (IoU(bb^{(kci)}, bb^{(kcj)}) > t)
\end{equation}
This gives $N_{IoU@0.7} = 715$ and $N_{IoU@0.5} = 5179$.

\paragraph{Missed objects in anchor based detector.}
RetinaNet~\cite{lin2018focal} assigns anchors to a ground truth bounding box if they have $> 0.5$ IoU.
In the case that a ground truth bounding box has not been covered by any anchor with IoU $>0.5$, the anchor with the largest IoU will be assigned to it. 
We calculate how often this forced assignment happens.
We use $15$ anchors ($5$ size: 32, 64, 128, 256, 512, and $3$ aspect-ratio: 0.5, 1, 2,  as is in RetinaNet~\cite{lin2018focal}) at stride $S = 16$.
For each image, after resizing it as its shorter edge to be $800$~\cite{lin2018focal}, we place these anchors at positions $\{(S / 2 + i \times S, S / 2 + j \times S)\}$, where $i \in [0, \lfloor\frac{(W - S/2)}{S}\rfloor]$ and $j \in [0, \lfloor\frac{(H - S/2)}{S}\rfloor]$. W, H are the image weight and height (the smaller one is equal to 800).
This results in a set of anchors $\mathcal{A}$. $|\mathcal{A}| = 15 \times \lfloor\frac{(W - S/2)}{S} + 1\rfloor \times \lfloor\frac{(H - S/2)}{S} + 1\rfloor$.
We calculate the number of the forced assignments by:
\begin{equation}
    N_{anchor} = \sum_{k=1}^{N} \sum_{i=1}^{n^{(k)}} \mathbbm{1} ((\max_{A \in \mathcal{A}}  IoU(bb^{(k\cdot i)}, A)) < 0.5)
\end{equation}
RenitaNet requires $N_{anchor} = 170220$ forced assignments: $125831$ for small objects ($35.3\%$ of all small objects), $18505$ for medium objects ($6.3\%$ of all medium objects), and $25884$ for large objects ($12.4\%$ of all large objects).

\section*{Appendix D: Experiments on PascalVOC}

\begin{table}[t]
\centering
\begin{tabular}{l c c c}
\hline
 & Resolution  & mAP$@$0.5 & FPS \\
 \hline
 Faster RCNN~\cite{ren2015faster} & $600 \times 1000$ & 76.4 & 5 \\
 Faster RCNN*~\cite{chen17implementation} &  $600 \times 1000$ & 79.8 & 5 \\
R-FCN~\cite{dai2016r} &  $600 \times 1000$ & 80.5 & 9 \\
 Yolov2~\cite{redmon2017yolo9000} & $544 \times 544$ & 78.6 & 40 \\
 SSD~\cite{fu2017dssd} & $513 \times 513$ & 78.9 & 19 \\
 DSSD~\cite{fu2017dssd} & $513 \times 513$ & 81.5 & 5.5 \\
 RefineDet~\cite{zhang2018single} & $512 \times 512$ & 81.8 & 24 \\
\hline
CenterNet-Res18 & $384 \times 384$ & 72.6 & 142\\
CenterNet-Res18 & $512 \times 512$ & 75.7 &  100\\
CenterNet-Res101 &$384 \times 384$ & 77.6 & 45\\ 
CenterNet-Res101 & $512 \times 512$  & 78.7 & 30\\
CenterNet-DLA  & $384 \times 384$& 79.3 & 50\\
CenterNet-DLA & $512 \times 512$  & 80.7& 33\\
\hline
\end{tabular}
\caption{Experimental results on Pascal VOC 2007 test. The results are shown in mAP$@0.5$. Flip test is used for CenterNet. The FPSs for other methods are copied from the original publications.}
\label{tab:pascal}
\end{table}

Pascal VOC~\cite{pascal-voc-2012} is a popular small object detection dataset. 
We train on VOC 2007 and VOC 2012 trainval sets, and test on VOC 2007 test set.
It contains $16551$ training images and $4962$ testing images of $20$ categories. 
The evaluation metric is mean average precision (mAP) at IOU threshold $0.5$.

We experiment with our modified ResNet-18, ResNet-101, and DLA-34 (See main paper Section. 5) in two training resolution: $384 \times 384$ and $512 \times 512$.
For all networks, we train $70$ epochs with learning rate dropped $10 \times$ at $45$ and $60$ epochs, respectively.
We use batchsize $32$ and learning rate $1.25e$-4 following the linear learning rate rule~\cite{goyal2017accurate}.
It takes one GPU $7$ hours/ $10$ hours to train in $384 \times 384$ for ResNet-101 and DLA-34, respectively.
And for $512 \times 512$, the training takes the same time in two GPUs.
Flip augmentation is used in testing.
All other hyper-parameters are the same as the COCO experiments.
We do not use Hourglass-104~\cite{Law_2018_ECCV} because it fails to converge in a reasonable time (2 days) when trained from scratch.

The results are shown in Table.~\ref{tab:pascal}.
Our best CenterNet-DLA model performs competitively with top-tier methods, and keeps a real-time speed.

\section*{Appendix E: Error Analysis}

\begin{table}
\centering
\begin{tabular}{l c c c}
\hline
 & AP & $AP_{50}$ & $AP_{75}$ \\
\hline
          & 36.3 & 54.0 & 39.6 \\
w/ gt size & 41.9 & 56.6 & 45.4 \\
w/ gt heatmap & 54.2 & 82.6 & 58.1 \\
w/ gt heatmap+size & 83.1 & 97.9 & 90.1 \\
w/ gt hm.+size+offset & 99.5 & 99.7 & 99.6 \\
\hline
\end{tabular}
\caption{Error analysis on COCO validation. We show COCO AP($\%$) after replacing each network prediction with its ground truth.}
\label{tab:gt}
\end{table}

We perform an error analysis by replacing each output head with its ground truth.
For the center point heatmap, we use the rendered Gaussian ground truth heatmap.
For the bounding box size, we use the nearest ground truth size for each detection.

The results in Table~\ref{tab:gt} show that improving both size map leads to a modest performance gain, while the center map gains are much larger.
If only the keypoint offset is not predicted, the maximum AP reaches $83.1$.
The entire pipeline on ground truth misses about $0.5\%$ of objects, due to discretization and estimation errors in the Gaussian heatmap rendering.

\end{document}